\PassOptionsToPackage{unicode}{hyperref}
\PassOptionsToPackage{hyphens}{url}
\documentclass[
]{article}
\usepackage{xurl}
\usepackage{xcolor}
\usepackage{amsmath,amssymb}
\usepackage{iftex}
\ifPDFTeX
  \usepackage[T1]{fontenc}
  \usepackage[utf8]{inputenc}
  \usepackage{textcomp} 
\else 
  \usepackage{unicode-math} 
  \defaultfontfeatures{Scale=MatchLowercase}
  \defaultfontfeatures[\rmfamily]{Ligatures=TeX,Scale=1}
\fi
\usepackage{lmodern}
\ifPDFTeX\else
\fi
\IfFileExists{upquote.sty}{\usepackage{upquote}}{}
\IfFileExists{microtype.sty}{
  \usepackage[]{microtype}
  \UseMicrotypeSet[protrusion]{basicmath} 
}{}
\makeatletter
\@ifundefined{KOMAClassName}{
  \IfFileExists{parskip.sty}{%
    \usepackage{parskip}
  }{
    \setlength{\parindent}{0pt}
    \setlength{\parskip}{6pt plus 2pt minus 1pt}}
}{
  \KOMAoptions{parskip=half}}
\makeatother
\usepackage{longtable,booktabs,array}
\usepackage{calc} 
\usepackage{etoolbox}
\makeatletter
\patchcmd\longtable{\par}{\if@noskipsec\mbox{}\fi\par}{}{}
\makeatother
\IfFileExists{footnotehyper.sty}{\usepackage{footnotehyper}}{\usepackage{footnote}}
\makesavenoteenv{longtable}
\ifLuaTeX
  \usepackage{luacolor}
  \usepackage[soul]{lua-ul}
\else
  \usepackage{soul}
\fi
\usepackage{bookmark}
\IfFileExists{xurl.sty}{\usepackage{xurl}}{} 
\hypersetup{
  hidelinks,
  pdfcreator={LaTeX via pandoc}}

\author{}
\date{}
\usepackage{needspace}
\begin{document}

\section{\textbf{The ÌròyìnSpeech Text Corpus: 24,905 Curated Yorùbá Sentences
for Speech and Language Technology}}

\textbf{Kọ́lá Túbọ̀sún}¹, \textbf{Anuoluwapo Aremu}¹, \textbf{Tolúlọpẹ́
Ògúnrẹ̀mí}², \textbf{Iroro Orife}³, \textbf{David Ifeoluwa Adelani}\textsuperscript{4}

¹ YorubaName.com · ² Stanford University · ³ Niger-Volta LTI · \textsuperscript{4} Mila /
McGill University

\section{}\label{section}

\section{\texorpdfstring{\textbf{Abstract}}{Abstract}}\label{abstract}

\emph{ÌròyìnSpeech is a 42-hour, 80-speaker Yorùbá read-speech corpus whose audio has been distributed by ELRA since 2024. This paper describes the release of its text component: 24,905 unique, hand-verified, tone-marked Yorùbá sentences (275,897 tokens; 15,686 types), curated in 2022 as recording prompts. Roughly 11,000 sentences were adapted from openly licensed news material; the remainder were written in-house to broaden coverage beyond the religious translation that dominates existing Yorùbá corpora. Every sentence was checked by hand for tone-mark accuracy, edited for read-aloud clarity and a neutral register, and localised so that non-Yorùbá personal and place names appear in Yorùbá form. Preparing the text for release surfaced systematic Unicode normalisation failures affecting more than 60\% of lines (with precomposed and decomposed forms of the same letter co-occurring within single sentences), which we document and correct. The corpus serves as a validated prompt set for new recording, and as tone-marked text for diacritic restoration, orthographic research, and TTS front-end and rule-based grapheme-to-phoneme work\textbf{.}}

\textbf{Keywords:} Yorùbá; low-resource languages; speech corpora;
diacritics; Unicode normalisation; African language technology; text corpus; tone marking

\section{\texorpdfstring{\textbf{1.
Overview}}{1. Overview}}\label{overview}

\subsection{\texorpdfstring{\textbf{1.1
Context}}{1.1 Context}}\label{context}

Yorùbá is spoken by more than forty million people across south-western
Nigeria, the Republic of Benin, Togo and a substantial global diaspora,
and yet the text available for building language technology in it
remains dominated by religious translation. Bible and
Jehovah\textquotesingle s Witness material make up the overwhelming bulk
of the Yorùbá side of the standard multilingual corpora (Agić \& Vulić, 2019), with the predictable consequence that models trained on them handle scripture
fluently and contemporary speech poorly. The same pattern holds for speech: BibleTTS (Meyer et al., 2022), one of the larger African speech resources, is built from Bible recordings, and earlier open Yorùbá speech data (Gutkin et al., 2020) is limited in size and domain.

The ÌròyìnSpeech project was undertaken in 2022, under an Imminent
Research Grant from \emph{Translated}, to address this at the level of
the source material rather than the model. Its aim was a modern,
secular, multi-purpose Yorùbá speech corpus, and it produced 42 hours of
studio audio recorded by 80 volunteers, described in Ògúnrẹ̀mí et al.
(2024) and distributed by ELRA as ELRA-S0492. 

Building that corpus required first building a text corpus: some 25,000
sentences, curated, corrected, tone-marked and shaped over several
months into prompts a volunteer could read and record in a studio. This
paper releases and describes the text for the first time.

The importance of this material rests on the need for a curated,
fully-diacritised, secular Yorùbá text collection which remains scarce, since most Yorùbá written online omits diacritics for lack of input tools (Túbọ̀sún, 2024). Still,  its most valuable applications
are not always related to speech. Diacritic restoration in Yorùbá,
tone-marking evaluation, grapheme-to-phoneme development and
orthographic research, for instance, all need this contemporary text
written and vetted by humans.

A narrative account of the project's design, recording and post-production is given in Túbọ̀sún (2024). It was written before post-production was complete, so its approximate figures of fifty hours and one hundred volunteers are superseded by those reported here, which exclude recordings that failed verification.

\subsection{\texorpdfstring{\textbf{1.2 Spatial
coverage}}{1.2 Spatial coverage}}\label{spatial-coverage}

The work mainly covers south-western Nigeria, where the language is
spoken as L1. Source news material concerns Nigerian national and
regional affairs; the localisation policy applied during curation (see
Method) means that place names appearing in the text skew toward
Yorùbá-speaking states. Although Yorùbá is also spoken in Benin and
Togo, those varieties are not represented; curation targeted the
Standard Yorùbá of Nigerian broadcast media.

\subsection{\texorpdfstring{\textbf{1.3 Temporal
coverage}}{1.3 Temporal coverage}}\label{temporal-coverage}

Source material dates from approximately 2018--2022. Curation was
carried out in 2022.

\section{\texorpdfstring{\textbf{2. Method}}{2. Method}}\label{method}

\subsection{\texorpdfstring{\textbf{2.1 Steps}}{2.1 Steps}}\label{steps}

Roughly 11,000 sentences were drawn from two openly-licensed
collections: MENYO-20k (Adelani et al., 2021), whose Yorùbá side derives
principally from Voice of Nigeria and Global Voices, and the Yorùbá
portion of MasakhaNER 2.0 (Adelani et al., 2022), which includes text from the Asejere newspaper. Both had already been preliminarily checked by native
speakers for the diacritic errors endemic to crawled Yorùbá web text,
which is why they were preferred over crawling afresh.

That was not enough for forty hours of speech. The remaining sentences,
about 12,000, were generated in-house. Curators worked from unpublished
Yorùbá short stories translated by Túbọ̀sún, splitting long sentences,
and writing new sentences on themes and in registers suggested by the
seed material. A further small set of sentences was composed to raise
the frequency of the rarer syllabic nasals. The syllabic \emph{m} (as in
\emph{m̀bá}, \emph{ḿbọ̀}) and nasals internal to proper names (as in
\emph{Bím̀bọ́}, \emph{Olúyọḿbọ̀}) were generated, to fill a gap that a
naturalistic sample leaves; a few dozen such sentences entered the
released corpus.

Beyond the syllabic nasals, coverage was not engineered to any other phonetic target, since the naturalistic text already supplied it. All 25 letters of the Yorùbá alphabet occur in the corpus, the rarest being \textit{h} (2,784 occurrences). Each of the seven oral vowels occurs with each of the three tones, the rarest combination being mid-tone \textit{u} (9,254). All 126 consonant–vowel combinations occur, as do 376 of the 378 tone-bearing CV syllables; the two absent are \textit{hè} and \textit{nè}. Tone-marked syllabic nasals occur 6,559 times (\textit{ń} 5,017; \textit{ǹ} 1,185; \textit{m̀} 221; \textit{ḿ} 136).

Every sentence, whether sourced or written, was then passed through a
manual editorial process:

\begin{itemize}
\item
  verifying tone marks and sub-dots word by word;
\item
  shortening and simplifying sentences where length or syntax would trip
  a reader;
\item
  revising religiously divisive and offensive terms toward a neutral
  register;
\item
  localising non-Yorùbá personal and place names into Yorùbá. e.g.
  United States to Ìlú Ọba, Kaduna to Ọ̀yọ́.
\end{itemize}

That last policy was adopted so that speakers would not be forced to
code-switch mid-utterance into English (or any other language)
phonology, which corrupts recordings intended for acoustic modelling. It
also means the corpus is not a faithful record of the news events its
source material reported. It is Yorùbá prose about a lightly
fictionalised Nigeria.

\subsection{\texorpdfstring{\textbf{2.2 Sampling
strategy}}{2.2 Sampling strategy}}\label{sampling-strategy}

The corpus is the complete prompt set as it stood at the close of
recording in 2022, including the sentences ultimately recorded on
Mozilla Common Voice as well as those recorded in-house. Sentences contributed to Common Voice (Ardila et al., 2020) are released there into the public domain under CC0. A record of which sentences were contributed is maintained in the project repository (\href{https://github.com/Niger-Volta-LTI/yoruba-voice}{https://github.com/Niger-Volta-LTI/yoruba-voice}); a future version of this release may flag them directly.

\subsection{\texorpdfstring{\textbf{2.3 Quality
control}}{2.3 Quality control}}\label{quality-control}

Original quality control was editorial and manual, as described above. 

During the 2022 project, the sentences were divided into batches among several editors, and each edited batch was then reviewed by the first author, who gave final approval. Corrections were made directly in the working spreadsheets and were not logged. Because each sentence was edited in sequence rather than independently annotated by more than one person, neither a direct error count nor inter-annotator agreement can be reported. An indirect estimate is possible for the sourced material, by aligning each sentence with its closest MENYO-20k or MasakhaNER 2.0 counterpart. Diacritics are ignored for the alignment, and the threshold is a string similarity of 0.6, looser than the provenance estimate in Section 2.4. Of the 1,112 sentences aligned this way, 379 (34\%) differ from their source in tone marks or sub-dots. Encoding differences such as U+0329 for U+0323 are set aside. Of 11,374 aligned words, 95.5\% are identical, 3.5\% differ in tone marking and 1.2\% in sub-dots. The rate differs by source (Table 1). Tone changes are most frequent in material from Asejere, which was annotated for named entities and not primarily for diacritics. Many differences are corrections of the kind the editorial pass targeted (\textit{Nàíjíríà} → \textit{Nàìjíríà}, \textit{Buhari} → \textit{Bùhárí}, \textit{se} → \textit{ṣe}); others reflect rewording that changed the word itself  (\textit{ni} $\leftrightarrow$ \textit{ní}), and some are likely errors introduced during curation. The comparison measures disagreement between the two versions, not which is correct. The word-level agreement between two independent annotation passes, that of the upstream projects and our own, is therefore about 95\%.

\begin{table}[htbp]
\centering
\small
\caption{Tone-mark and sub-dot differences between aligned sentences and their upstream sources. Asejere sentences are from MasakhaNER 2.0; Voice of Nigeria and Global Voices sentences are from MENYO-20k. Sentences found in both corpora are counted as Voice of Nigeria.}
\begin{tabular}{lrrrrr}
\toprule
Source & \shortstack[r]{Sentences\\aligned} & \shortstack[r]{Sentences\\changed} & \shortstack[r]{Aligned\\words} & \shortstack[r]{Tone\\differs} & \shortstack[r]{Sub-dot\\differs} \\
\midrule
Asejere & 294 & 118 & 2,579 & 5.8\% & 0.5\% \\
Voice of Nigeria & 768 & 248 & 8,179 & 3.0\% & 1.4\% \\
Global Voices & 50 & 13 & 616 & 1.5\% & 1.1\% \\
\midrule
\textbf{All} & \textbf{1,112} & \textbf{379} & \textbf{11,374} & \textbf{3.5\%} & \textbf{1.2\%} \\
\bottomrule
\end{tabular}
\end{table}
 
A
second pass was carried out for this release, and found problems that
hand-checking in a spreadsheet did not catch. The most significant is Unicode normalisation. 15,451 of the 25,001
lines were not in NFC form. They were, instead, precomposed and
decomposed representations of the same letter, frequently co-occurring
within a single sentence. In one instance, \emph{à} is precomposed as
U+00E0, whereas a neighbouring vowel \emph{ì} is encoded as \emph{i}
followed by U+0300. The drift is confined entirely to characters that do
possess a precomposed form and were simply not using it: \emph{í},
\emph{à}, \emph{á}, \emph{ì}, \emph{ó}, \emph{ú}, \emph{é}, \emph{ò},
\emph{è}, \emph{ù}, \emph{ń} and \emph{ọ} --- with \emph{í} alone
accounting for 25,158 occurrences.

Thirty-five sentences used U+0329 COMBINING VERTICAL LINE BELOW in place
of U+0323 COMBINING DOT BELOW. To the human viewer, the letters look the
same, but the difference is consequential downstream. Two encodings of
the same word will not match, will not sort together, will not tokenise
identically, and will inflate vocabulary counts. The problem has been
fixed, and the corpus is now uniformly NFC, with the raw file retained
for reproducibility.

Normalising to NFC does not, however, make the corpus uniformly
precomposed. 22,642 of the 24,905 sentences (90.9\%, across 69,038
occurrences) contain at least one instance of \emph{ọ́}, \emph{ọ̀},
\emph{ẹ́} or \emph{ẹ̀}. These have no precomposed form in Unicode: none was ever encoded, and under Unicode's normalisation stability policy none can now be added (Túbọ̀sún, 2026a). Each therefore remains two code points after normalisation. Any downstream tool that assumes NFC yields one code point per character will miscount characters, which affects character error rates and character-level models, and may separate a letter from its marks when segmenting text. Exact matching is unaffected, provided both texts are normalised to the same form.

Also corrected: 96 exact duplicates, 5 no-break spaces, inconsistent
quotation marks in 505 sentences, and stray editorial annotations that
had become attached to five sentences. Eight sentences of fewer than
four words, seven containing Igbo characters and five containing letters
outside the Yorùbá alphabet are flagged rather than removed. A further 14 sentences contain a combining tone mark that is not attached to a letter (following a space, punctuation or the start of the line), and one uses U+00B8 SPACING CEDILLA in place of a comma. These curation errors are also flagged rather than changed.

Contour tones are written as they were in 2022: as geminated vowels, the convention common in modern Yorùbá writing, where the curators wrote them out, and otherwise unmarked. The caron and circumflex convention proposed in Túbọ̀sún (2026b) has not been applied. Users who wish to apply that convention can do so through its equivalence mappings, which convert between geminated and contour-marked forms.

\subsection{\texorpdfstring{\textbf{2.4
Limitations}}{2.4 Limitations}}\label{limitations}

\textbf{Provenance cannot be fully reconstructed.} Matching the corpus
against the published MENYO-20k and MasakhaNER 2.0 releases recovers
only 65 exact matches and 921 further sentences with substantial n-gram
overlap, far short of the \textasciitilde11,000 the
project\textquoteright s records indicate were sourced. This is because
of the editorial processes described earlier: splitting, shortening,
name-localisation and other edits that leave a derived sentence
unrecognisable at the string level. A per-sentence provenance estimate
is supplied, with explicit confidence bands, but it is a reconstruction
and should not be read as a licensing determination.

\textbf{Licence.} Both upstream corpora carry non-commercial terms.
Because the derived material cannot now be cleanly separated, the whole
corpus is released CC BY-NC 4.0. Licence incompatibility of this kind is common and often unexamined in African-language corpus releases (van Gassen, 2026); we have adopted the most restrictive upstream terms rather than risk misrepresenting them.\footnote{The 2024 paper (Ògúnrẹ̀mí et al., 2024) listed the licence as CC BY 4.0. The text released here is licensed CC BY-NC 4.0, because it is derived in part from upstream material carrying non-commercial terms that can no longer be cleanly separated from the in-house sentences.} This licence cannot apply to any sentence that is also available through Mozilla Common Voice under CC0 (see Section 2.2); users requiring unrestricted terms for specific sentences can identify them from the record described in Section 2.2.

\textbf{Register.} Read-aloud news and adapted fiction. Neither
conversational Yorùbá nor the full range of written registers is
represented, and the localisation policy makes the corpus unsuitable as
a record of factual events.

\textbf{Dialect.} Curation targeted Standard Yorùbá as used in broadcast
media. Regional variation is not represented.

\textbf{Residual errors.} Despite manual verification, the corpus still contains errors in tone marking, sub-dots, spelling and hyphenation, as well as variant spellings of the same word (for example, \textit{nàìjíríà} and \textit{nàìjííríà}); several surfaced while compiling a word list from this release. No further correction pass has been made for this version. Corrections reported by users through the project repository (\href{https://github.com/Niger-Volta-LTI/yoruba-voice}{https://github.com/Niger-Volta-LTI/yoruba-voice}) will be incorporated into future versions of the release.

\subsection{\texorpdfstring{\textbf{2.5 Use of AI tools}}{2.5 Use of AI tools}}

The analyses reported in Sections 2.1, 2.3 and 4 were carried out with the assistance of Claude (Anthropic), an AI assistant, which wrote and ran the analysis scripts under the first author’s direction. The authors reviewed all methods, results and text, and take full responsibility for them. Scripts for the analyses in Sections 2.1, 2.3 and 4 are available at https://github.com/Niger-Volta-LTI/yoruba-voice.

\newpage
\section{\texorpdfstring{\textbf{3. Dataset
description}}{3. Dataset description}}\label{dataset-description}

{\def\LTcaptype{none} 
\begin{center}
\begin{tabular}{@{}l>{\raggedright\arraybackslash}p{0.62\linewidth}@{}}
\toprule
\textbf{Field} & \textbf{Value} \\
\midrule
Object name & ÌròyìnSpeech Text Corpus \\
Format & UTF-8 plain text (.txt), CSV, JSON Lines \\
Creation dates & 2022-01 to 2022-12 (curation); 2026 (cleaning and packaging) \\
Dataset creators & Kọ́lá Túbọ̀sún, Anuoluwapo Aremu, Tolúlọpẹ́ Ògúnrẹ̀mí, Iroro Orife, David Ifeoluwa Adelani \\
Language & Yorùbá (ISO 639-3: yor) \\
Licence & CC BY-NC 4.0 \\
Repository & Zenodo, doi:10.5281/zenodo.23138464 \\
Related resource & ELRA-S0492 (audio), ISLRN 012-405-700-001-6 \\
Publication date & 2026 \\
\bottomrule
\end{tabular}
\end{center}
}
\needspace{12\baselineskip}
\subsection{\texorpdfstring{\textbf{Corpus
statistics}}{Corpus statistics}}\label{corpus-statistics}

{\def\LTcaptype{none} 
\begin{center}
\begin{tabular}{@{}ll@{}}
\toprule
\textbf{Measure} & \textbf{Value} \\
\midrule
Rows in the 2022 working file & 25,001 \\
Unique sentences after cleaning & 24,905 \\
Exact duplicates removed & 96 \\
Word tokens & 275,897 \\
Word types & 15,686 \\
Hapax legomena & 8,444 \\
Characters (Unicode code points, NFC) & 1,344,571 \\
Characters (grapheme clusters) & 1,275,178 \\
Mean sentence length & 11.1 words (range 2--26) \\
Distinct code points & 113 \\
\bottomrule
\end{tabular}
\end{center}
}

\section{\texorpdfstring{\textbf{4. Reuse
potential}}{4. Reuse potential}}\label{reuse-potential}

The obvious reuse of this work is as a prompt set. The sentences are
already validated for read-aloud recording, and a group wanting to build
a new Yorùbá voice, either a different dialect, an under-recorded age
group, a clinical or accented population or a different gender, can
begin recording immediately rather than spending months on curation, as
we did.

There are other obvious re-uses like automatic diacritic restoration (ADR). ADR is the central preprocessing problem in Yorùbá
NLP, since most Yorùbá text in circulation is untoned or partially
toned, and progress on it is bounded by the availability of
correctly-toned text to train and evaluate against. This corpus supplies
275,897 tokens of it, hand-verified, in a contemporary secular register.
Because Standard Yorùbá orthography is close to phonemic, with tone marked on every syllable, a pronunciation can be derived from this text by rule, so the corpus can serve as input for building and evaluating grapheme-to-phoneme systems and TTS front ends. The release itself contains no phonemic transcriptions. The corpus is also usable for tone-marking evaluation, and as a clean-text counterweight in language-model training sets otherwise
dominated by scripture.

To illustrate this use, we trained a simple ADR baseline on this corpus and, separately, on the Yorùbá side of the MENYO-20k training set. The baseline is a bigram word model decoded with the Viterbi algorithm, which restores each undiacritised word to a diacritised form seen in training. Each model was tested on held-out sentences from both corpora (Table 2). Sentences close to the MENYO-20k test and development sets were removed from our training data. Our test set is 2,500 sentences written in-house. Trained on the same amount of text (about 200,000 tokens), the two corpora perform almost identically on MENYO-20k's own test set (84.2\% and 84.7\% word accuracy). The model trained on MENYO-20k, however, falls to 78.1\% on ÌròyìnSpeech text, against 86.5\% for the in-domain model. Combining the two raises accuracy on the MENYO-20k test set to 87.0\%. The baseline is deliberately simple; its purpose is to show that the corpus complements existing Yorùbá data, not to set a state of the art. Because our test sentences were written by the same curators as the training data, in-domain scores are probably flattered.

\begin{table}[htbp]
\centering
\small
\caption{Word-level accuracy (\%) of a bigram ADR baseline}
\begin{tabular}{p{4.6cm}rrr}
\toprule
Training data & Tokens & \shortstack[r]{Test:\\ÌròyìnSpeech} & \shortstack[r]{Test:\\MENYO-20k} \\
\midrule
ÌròyìnSpeech (full) & 239,123 & 86.9 & 84.5 \\
ÌròyìnSpeech (size-matched) & 201,688 & 86.5 & 84.2 \\
MENYO-20k train & 201,683 & 78.1 & 84.7 \\
Both & 440,806 & \textbf{87.2} & \textbf{87.0} \\
\bottomrule
\end{tabular}
\end{table}

For research on Yorùbá orthography specifically, the raw file is a
well-documented record of how encoding inconsistency arises despite
months of careful manual work with tools that give no indication that
two identical-looking characters are not the same character. The failure
mode it exhibits is general to every language whose orthography requires
combining marks. The corpus, which has been tone-marked and
hand-verified, is also a suitable basis for future research into other
aspects of tonal annotation, including the representation of contour
tones, which standard Yorùbá orthography leaves unmarked, and for which
Túbọ̀sún (2026b) provides a recommendation that has not been applied to this release (see Section 2.3).

Finally, the text corresponds to the 42 hours of audio released as ELRA-S0492. The changes made for this release are confined to encoding, punctuation and the removal of duplicates and stray annotations, so the text still corresponds to the recordings. Researchers who license ELRA-S0492 can now work with the text under a
separate, more permissive access regime --- a division that may be
useful for text-only work that does not warrant a corpus licence.

\section{\texorpdfstring{\textbf{Acknowledgements}}{Acknowledgements}}\label{acknowledgements}

\emph{Volunteers, Imminent/Translated, University of Lagos Department of
Linguistics.}

\section{\texorpdfstring{\textbf{Funding
Statement}}{Funding Statement}}\label{funding-statement}

Curation and recording were supported by a 2022 Imminent Research Grant
from Translated.

\section{\texorpdfstring{\textbf{Competing
Interests}}{Competing Interests}}\label{competing-interests}

The authors are also authors of the ÌròyìnSpeech corpus (Ògúnrẹ̀mí et
al., 2024) that this dataset derives from.

\section{\texorpdfstring{\textbf{Data Accessibility Statement}}{Data Accessibility Statement}}\label{data-accessibility-statement}

The ÌròyìnSpeech Text Corpus is openly available under a CC BY-NC 4.0 licence. The corresponding audio is distributed by ELRA as ELRA-S0492 (ISLRN~\mbox{012-405-700-001-6}). The corpus is deposited on Zenodo:\newline
\mbox{\url{https://doi.org/10.5281/zenodo.23138464}}.

\section{\texorpdfstring{\textbf{Ethics and Consent}}{Ethics and Consent}}\label{ethics-and-consent}

The corpus consists of sentences adapted from openly licensed news material and sentences written by the project team; no personal data were collected from participants for this release. The recording of the associated audio by volunteer speakers, under informed consent, is described in Ògúnrẹ̀mí et al. (2024); no audio or speaker information is included here.

\section{\texorpdfstring{\textbf{Author Contributions}}{Author Contributions}}\label{author-contributions}

Kọ́lá Túbọ̀sún: project administration, conceptualisation, data curation, cleaning and analysis for this release, writing (original draft). Anuoluwapo Aremu: data curation. Tolúlọpẹ́ Ògúnrẹ̀mí: data curation, writing (review and editing). Iroro Orife: software (recording application and validation scripts), data curation (dataset processing, metadata consistency), writing (review and editing).
David Ifeoluwa Adelani: project administration, conceptualisation, resources (source corpora), writing (review and editing).

\section{}\label{section-1}

\section{\texorpdfstring{\textbf{References}}{References}}\label{references}

\raggedright

Adelani, D. I., Neubig, G., Ruder, S., Rijhwani, S., Beukman, M.,
Palen-Michel, C., Lignos, C., Alabi, J. O., Muhammad, S. H., Nabende,
P., Dione, C. M. B., Bukula, A., Mabuya, R., Dossou, B. F. P., Sibanda,
B., Buzaaba, H., Mukiibi, J., Kalipe, G., Mbaye, D., \ldots{} Klakow, D.
(2022). MasakhaNER 2.0: Africa-centric transfer learning for named
entity recognition. In \emph{Proceedings of the 2022 Conference on
Empirical Methods in Natural Language Processing} (pp. 4488--4508).
Association for Computational Linguistics.
\url{https://doi.org/10.18653/v1/2022.emnlp-main.298}

Adelani, D. I., Ruiter, D., Alabi, J. O., Adebonojo, D., Ayeni, A.,
Adeyemi, M., Awokoya, A. E., \& España-Bonet, C. (2021). The effect of
domain and diacritics in Yoruba--English neural machine translation. In
K. Duh \& F. Guzmán (Eds.), \emph{Proceedings of Machine Translation
Summit XVIII: Research Track} (pp. 61--75). Association for Machine
Translation in the Americas.
\url{https://aclanthology.org/2021.mtsummit-research.6/}

Agić, Ž., \& Vulić, I. (2019). JW300: A wide-coverage parallel corpus
for low-resource languages. In \emph{Proceedings of the 57th Annual
Meeting of the Association for Computational Linguistics} (pp.
3204--3210). \url{https://doi.org/10.18653/v1/P19-1310}

Ardila, R., Branson, M., Davis, K., Kohler, M., Meyer, J., Henretty, M.,
Morais, R., Saunders, L., Tyers, F., \& Weber, G. (2020). Common Voice:
A massively-multilingual speech corpus. In \emph{Proceedings of the
Twelfth Language Resources and Evaluation Conference} (pp. 4218--4222).
European Language Resources Association.
\url{https://aclanthology.org/2020.lrec-1.520/}

Gutkin, A., Demirşahin, I., Kjartansson, O., Rivera, C., \& Túbọ̀sún, K.
(2020). Developing an open-source corpus of Yoruba speech. In
\emph{Interspeech 2020} (pp. 404--408). ISCA.
\url{https://doi.org/10.21437/Interspeech.2020-1096}

Meyer, J., Adelani, D. I., Casanova, E., Öktem, A., Whitenack, D.,
Weber, J., Kabongo Kabenamualu, S., Salesky, E., Orife, I., Leong, C.,
Ogayo, P., Emezue, C. C., Mukiibi, J., Osei, S., Agbolo, A., Akinode,
V., Opoku, B., Samuel, O., Alabi, J., \& Muhammad, S. H. (2022).
BibleTTS: A large, high-fidelity, multilingual, and uniquely African
speech corpus. In \emph{Interspeech 2022} (pp. 2383--2387). ISCA.
\url{https://doi.org/10.21437/Interspeech.2022-10850}

Ògúnrẹ̀mí, T., Túbọ̀sún, K., Aremu, A., Orife, I., \& Adelani, D. I.
(2024). ÌròyìnSpeech: A multi-purpose Yorùbá speech corpus. In N.
Calzolari, M.-Y. Kan, V. Hoste, A. Lenci, S. Sakti, \& N. Xue (Eds.),
\emph{Proceedings of the 2024 Joint International Conference on
Computational Linguistics, Language Resources and Evaluation
(LREC-COLING 2024)} (pp. 9296--9303). ELRA and ICCL.
\url{https://aclanthology.org/2024.lrec-main.812/}

Túbọ̀sún, K. (2024, May 20). \emph{Making machines speak Yorùbá}.
Imminent. \url{https://imminent.translated.com/making-machines-speak-yoruba}

Túbọ̀sún, K. (2026a). \emph{Yorùbá in Unicode: An overview of a
problem}. arXiv. \url{https://arxiv.org/abs/2609.33734} (To appear in
\emph{Yorùbá print culture: A handbook}, Routledge)

Túbọ̀sún, K. (2026b). \emph{Marking contour tones in Yorùbá: A
typographic and computational proposal}. arXiv.
\url{https://arxiv.org/abs/2609.38627}

van Gassen, E. A. P. (2026). Open but unvetted: The ethics of African
language data. In M. Matfunjwa, M. Setaka, R. Mabuya, \& M. van Zaanen
(Eds.), \emph{Proceedings of Resources for African Indigenous Languages
(RAIL) 2026 @ LREC 2026} (pp. 128--139). ELRA.
\url{https://doi.org/10.63317/2ewucnr8r5hq}
\end{document}